\documentclass[10pt,twocolumn,letterpaper]{article}

\usepackage{titling}
\usepackage{iccv}
\usepackage{times}
\usepackage{epsfig}
\usepackage{graphicx}
\usepackage{amsmath}
\usepackage{amssymb}

\usepackage[utf8]{inputenc}
\usepackage{subcaption}
\usepackage{multirow}
\usepackage{pgfplots}
\usepackage{tikz}
\usepackage{tikzscale}
\usetikzlibrary{calc,trees,positioning,fit,arrows,decorations.pathreplacing}

\newcommand{\norm}[1]{\left\lVert #1 \right\rVert_2}

\newcommand{\invisnote}[1]{}

\usepackage[breaklinks=true,bookmarks=false]{hyperref}

\iccvfinalcopy % *** Uncomment this line for the final submission

\def\iccvPaperID{2245} % *** Enter the ICCV Paper ID here

\ificcvfinal\fi
\begin{document}

%%%%%%%%% TITLE
\title{Learning to Estimate 3D Hand Pose from Single RGB Images}

\author{Christian Zimmermann, Thomas Brox\\
University of Freiburg\\
{\tt\small \{zimmermann, brox\}@cs.uni-freiburg.de}
}

\maketitle
%\thispagestyle{empty}

% Input main document
%%%%%%%%% ABSTRACT
\begin{abstract}
Low-cost consumer depth cameras and deep learning have enabled reasonable 3D 
hand pose estimation from single depth images.
In this paper, we present an approach that estimates 3D hand pose from regular 
RGB images. This task has far more ambiguities due to the missing depth 
information. To this end, we propose a deep network that learns a 
network-implicit 3D articulation prior. Together with detected keypoints in the 
images, this network yields good estimates of the 3D pose. We introduce a large 
scale 3D hand pose dataset based on synthetic hand models for training the 
involved networks. Experiments on a variety of test sets, including one on sign 
language recognition, demonstrate the feasibility of 3D hand pose estimation on 
single color images. 
\end{abstract}

%%%%%%%%% BODY TEXT
%-------------------------------------------------------------------------
\section{Introduction}

The hand is the primary operating tool for humans. Therefore, its location, 
orientation and articulation in space is vital for many potential applications, 
for instance, object handover in robotics, learning from demonstration, sign 
language and gesture recognition, and using the hand as an input device for 
man-machine interaction. 

Full 3D hand pose estimation from single images is difficult because of many 
ambiguities, strong articulation, and heavy self-occlusion, even more so than 
for the overall human body. Therefore, specific sensing equipment like data 
gloves or markers are used, which restrict the application to limited scenarios. 
Also the use of multiple cameras severly limits the application domain. Most 
contemporary works rely on the depth image from a depth camera. However, depth 
cameras are not as commonly available as regular color cameras, and they only 
work reliably in indoor environments. 

\begin{figure*}
\centering
    \includegraphics{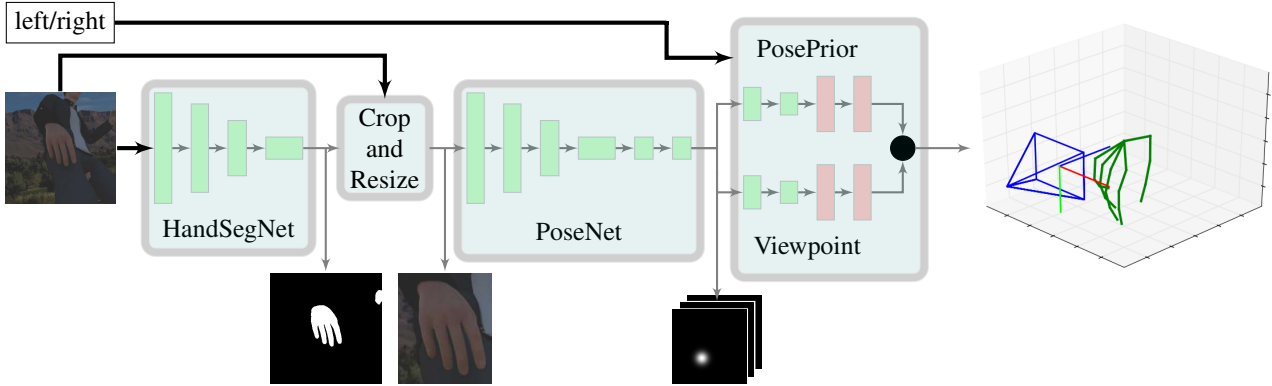}
\caption{Our approach consists of three building blocks. First, the hand is localized within the image by a segmentation network (\textit{HandSegNet}). Accordingly to the hand mask, the input image is cropped and serves as input to the \textit{PoseNet}. This localizes a set of hand keypoints represented as score maps $\mathbf{c}$. Subsequently, the \textit{PosePrior} network estimates the most likely 3D structure conditioned on the score maps. This figure serves for illustration of the overall approach and does not reflect the exact architecture of the individual building blocks.}\label{fig:overview}
\end{figure*}

In this paper, we present an approach to learn full 3D hand pose estimation from 
single color images without the need for any special equipment. We capitalize on 
the capability of deep networks to learn sensible priors from data in order to 
resolve ambiguities. Our overall approach consists of three deep networks that 
cover important subtasks on the way to the 3D pose; see 
Figure~\ref{fig:overview}. The first network provides a hand segmentation to 
localize the hand in the image. Based on its output, the second network 
localizes hand keypoints in the 2D images. The third network finally derives the 
3D hand pose from the 2D keypoints, and is the main contribution of this paper. 
In particular, we introduce a canonical pose representation to make this 
learning task feasible. 

Another difficulty compared to 3D pose estimation at the level of the human body 
is the restricted availability of data. While human body pose estimation can 
leverage several motion capture databases, there is hardly any such data for 
hands. To train a network, a large dataset with ground truth 3D keypoints is 
needed. Since there is no such dataset with sufficient variability, we created a 
synthetic dataset with various data augmentation options.

The resulting hand pose estimation system yields very promising results, both 
qualitatively and quantitatively on existing small-scale datasets. We also 
demonstrate the use of 3D hand pose for the task of sign language recognition. 
The dataset and our trained networks are available online. 
\footnote{\href{https://lmb.informatik.uni-freiburg.de/projects/hand3d/}{https://lmb.informatik.uni-freiburg.de/projects/hand3d/}}

\begin{figure}
\centering
\begin{subfigure}{.45\columnwidth}
  \centering
  \includegraphics[width=\linewidth]{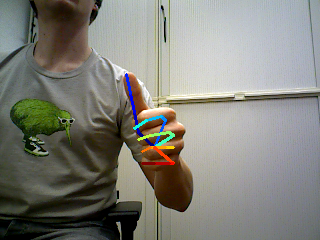}
\end{subfigure}
\begin{subfigure}{.5\columnwidth}
  \centering
  \includegraphics[width=\linewidth]{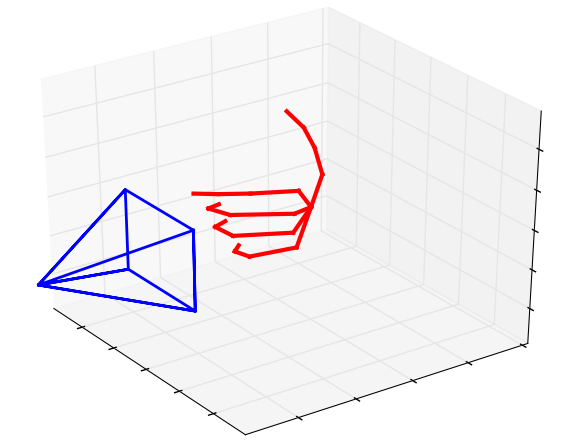}
\end{subfigure}
\caption{Given a color image we detect keypoints in 2D (shown overlayed) and 
learn a prior that allows us to estimate a normalized 3D hand pose.}
\label{fig:teaser}
\end{figure}

%-------------------------------------------------------------------------
\section{Related work}
\label{subsec:relatedWork}

\textbf{2D Human Pose Estimation.} Spurred by the MPII Human Pose benchmark \cite{andriluka_2d_2014} and the advent of Convolutional Neural Networks (CNN) this field made large progress in the last years. The CNN architecture of Toshev and Szegedy \cite{toshev_deeppose_2014} directly regresses 2D cartesian coordinates from color image input. More recent works like Thompson \etal~\cite{tompson_real-time_2014} and Wei \etal~\cite{wei_convolutional_2016} turned towards regressing score maps. For parts of our work, we employ a comparable network architecture as Wei \etal~\cite{wei_convolutional_2016}.

\textbf{3D Human Pose Estimation.} We only discuss the most relevant works here and refer to Sarafianos \etal~\cite{sarafianos_3d_2016} for more information.
Like our approach, many works use a two part pipeline \cite{tompson2014joint, chen2014articulated, chen_3d_2016, tome_lifting_2017, bogo_keep_2016}. First they detect keypoints in 2D to utilize the discriminative power of current CNN approaches and then attempt to lift the set of 2D detections into 3D space. Different methods for lifting the representation have been proposed: Chen \etal~\cite{chen_3d_2016} deployed a nearest neighbor matching of a given 2D prediction using a database of 2D to 3D correspondences. 
Tome \etal~\cite{tome_lifting_2017} created a probabilistic 3D pose model based upon a mixture of probabilistic PCA bases. 
Bogo \etal~\cite{bogo_keep_2016} optimizes the reprojection error between 3D joint locations of a statistical body shape model and 2D prediction.
Pavlakos \etal~\cite{pavlakos_coarse--fine_2016} proposed a volumetric approach that treats pose estimation as per voxel prediction of scores in a coarse-to-fine manner, which gives a natural representation to the data, but is computationally expensive and limited by the GPU memory to fit the voxel grid.
Recently, there have been several approaches that apply deep learning for lifting 2D keypoints to 3D pose for human body pose estimation \cite{zhao_simple_2016, moreno-noguer_3d_2016, popa_deep_2017}. Furthermore Mehta \etal~\cite{mehta_monocular_2016} uses transfer learning to infer the 3D body pose directly from images with a single network. 
While these works are all on 3D body pose estimation, we provide the first such work for 3D hand pose estimation, which is substantially harder due to stronger articulation and self-occlusion, as well as less data being available.

\textbf{Hand Pose Estimation.} 
Athitsos and Sclaroff \cite{athitsos_estimating_2003} proposed a single frame based detection approach based on edge maps and Chamfer matching. 
With the advent of low-cost consumer depth cameras, research focused on hand pose from depth data.
Oikonomidis \etal~\cite{oikonomidis_efficient_2011} proposed a technique based on Particle Swarm Optimization (PSO). 
Sharp \etal~\cite{sharp_accurate_2015} added the possibility for reinitialization. A certain number of candidate poses is created and scored against the observed depth image.
Tompson \etal~\cite{tompson_real-time_2014} used a CNN for detection of hand keypoints in 2D, which is conditioned on a multi-resolution image pyramid. The pose in 3D is recovered by solving an inverse kinematics optimization problem.
Approaches like Zhou \etal~\cite{zhou_model-based_2016} or Oberweger \etal~\cite{oberweger_hands_2015} train a CNN that directly regresses 3D coordinates given hand cropped depth maps. Whereas Oberweger \etal~\cite{oberweger_hands_2015} explored the possibility to encode correlations between keypoint coordinates in a compressing bottleneck, Zhou \etal~\cite{zhou_model-based_2016} estimate angles between bones of the kinematic chain instead of Cartesian coordinates.  
Oberweger \etal~\cite{oberweger_training_2015} utilizes a CNN that can synthesize a depth map from a given pose estimate. This allows them to successively refine initial pose estimates by minimizing the distance between the observed and the synthesized depth image. 

There aren't any approaches yet that tackle the problem of 3D hand pose estimation from a single color image with a learning based formulation. Previous approaches differ because they rely on depth data \cite{tompson_real-time_2014, zhou_model-based_2016, oberweger_hands_2015, oberweger_training_2015}, they use explicit models to infer pose by matching against a predefined database of poses \cite{athitsos_estimating_2003}, or they only perform tracking based on an initial pose rather than full pose estimation \cite{oikonomidis_efficient_2011, sharp_accurate_2015}.

%-------------------------------------------------------------------------

\begin{figure*}
\centering
    \includegraphics{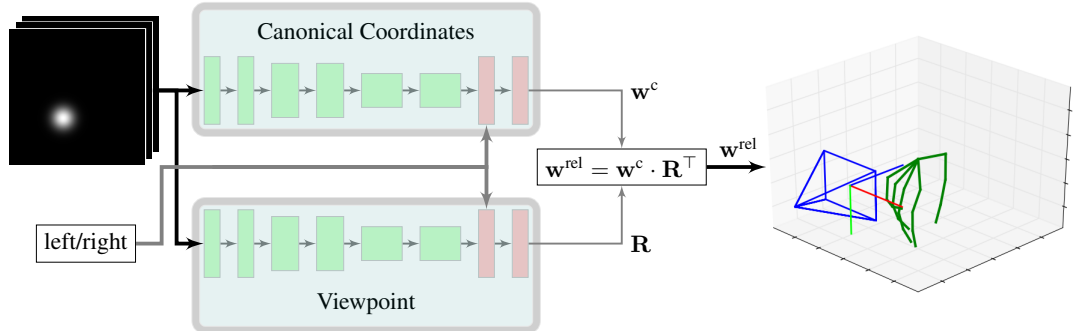}
\caption{Proposed architecture for the \textit{PosePrior} network. Two almost symmetric streams estimate canonical coordinates and the viewpoint relative to this coordinate system. Combination of the two predictions yields an estimation for the relative normalized coordinates $\mathbf{w}^\textit{rel}$.}\label{fig:arch_poseprior}
\end{figure*}

%-------------------------------------------------------------------------
\section{Hand pose representation}
\label{subsec:problem_definition}
Given a color image $I \in \mathbb{R}^{N\times M\times 3}$ showing a single hand, we want to infer its 3D pose. We define the hand pose by a set of coordinates $\mathbf{w}_i = (x_i, y_i, z_i)$, which describe the locations of $J$ keypoints in 3D space, i.e., $i \in [1, J]$ with $J=21$ in our case.  

The problem of inferring 3D coordinates from a single 2D observation is ill-posed. 
Among other ambiguities, there is a scale ambiguity. 
Thus, we infer a scale-invariant 3D structure by training a network to estimate normalized coordinates 
\begin{equation}
    \mathbf{w}^{\text{norm}}_{i} = \frac{1}{s} \cdot \mathbf{w}_i \text{,}
    \label{eq:scale_eq}
\end{equation}
where $s= \norm{\mathbf{w}_{k+1} - \mathbf{w}_{k}}$ is a sample dependent constant that normalizes the distance between a certain pair of keypoints to unit length.
We choose $k$ such that $s=1$ for the first bone of the index finger. 

Moreover, we use relative 3D coordinates to learn a translation invariant representation of hand poses. This is realized by subtracting the location of a defined root keypoint. The relative and normalized 3D coordinates are given by
\begin{equation}
    \mathbf{w}^{\text{rel}}_{i} = \mathbf{w}^{\text{norm}}_{i} - \mathbf{w}^{\text{norm}}_{r}
    \label{eq:rel_eq}
\end{equation} 
where $r$ is the root index. In experiments the palm keypoint was the most stable landmark. Thus we use $r=0$.

%-------------------------------------------------------------------------
\section{Estimation of 3D hand pose}
\label{subsec:technical_approach}

We estimate three-dimensional normalized coordinates $\mathbf{w}^{\text{rel}}$ from a single input image. An overview of the general approach is provided in Figure \ref{fig:overview}. In the following sections, we provide details on its components. 

\subsection{Hand segmentation with HandSegNet}
\label{subsubsec:hand_seg_net}
For hand segmentation we deploy a network architecture that is based on and initialized by the person detector of Wei \etal~\cite{wei_convolutional_2016}. They cast the problem of 2D person detection as estimating a score map for the center position of the human. The most likely location is used as center for a fixed size crop. Since the hand size drastically changes across images and depends much on the articulation, we rather cast the hand localization as a segmentation problem.  Our \textit{HandSegNet} is a smaller version of the network from Wei \etal~\cite{wei_convolutional_2016} trained on our hand pose dataset. Details on the network architecture and its training prcedure are provided in the supplemental material. The hand mask provided by \textit{HandSegNet} allows us to crop and normalize the inputs in size, which simplifies the learning task for the \textit{PoseNet}. 

\subsection{Keypoint score maps with PoseNet}
\label{subsubsec:pose_net}
We formulate localization of 2D keypoints as estimation of 2D score maps $\mathbf{c} = \{ \mathbf{c}_1(u, v), \hdots, \mathbf{c}_J(u, v)\}$. We train a network to predict $J$ score maps $\mathbf{c}_i  \in \mathbb{R}^{N\times M}$, where each map contains information about the likelihood that a certain keypoint is present at a spatial location.

The network uses an encoder-decoder architecture similar to the Pose Network by Wei \etal~\cite{wei_convolutional_2016}. Given the image feature representation produced by the encoder, an initial score map is predicted and is successively refined in resolution. We initialized with the weights from Wei \etal~\cite{wei_convolutional_2016}, where it applies, and retrained the network for hand keypoint detection. 
A complete overview over the network architecture is located in the supplemental material.

\subsection{3D hand pose with the PosePrior network}
\label{subsubsec:pose_prior}
The \textit{PosePrior} network learns to predict relative, normalized 3D coordinates conditioned on potentially incomplete or noisy score maps $\mathbf{c}(u, v)$. To this end, it must learn the manifold of possible hand articulations and their prior probabilities. Conditioned on the score maps, it will output the most likely 3D configuration given the 2D evidence.

Instead of training the network to predict absolute 3D coordinates, we rather propose to train the network to predict coordinates within a canonical frame and additionally estimate the transformation into the canonical frame.
Explicitly enforcing a representation that is invariant to the global orientation of the hand is beneficial to learn a prior, as we show in our experiments in section~\ref{subsec:lifting_2d_pose}. 
 
Given the relative normalized coordinates we propose to use a canonical frame $\mathbf{w}^\text{c}$, that relates to  $\mathbf{w}^\text{rel}$ in the following way: An intermediate representation
 \begin{equation}
    \mathbf{w}^\text{c*} = \mathbf{R}( \mathbf{w}^\text{rel} ) \cdot \mathbf{w}^\text{rel}
\end{equation}
with $\mathbf{R}(  \mathbf{w}^\text{rel} ) \in \mathbb{R}^{3 \times 3}$ being a 3D rotation matrix is calculated in a two step procedure. First, one seeks the rotation $\mathbf{R}_\text{xz}$ around the x- and z-axis such that a certain keypoint $\mathbf{w}^\text{c*}_a$ is aligned with the y-axis of the canonical frame:
\begin{equation}
    \mathbf{R}_\text{xz} \cdot \mathbf{w}^\text{c*}_a = \lambda \cdot (0, 1, 0)^\top\; \text{with $\lambda \geq 0$}.
\end{equation}
Afterwards, a rotation $\mathbf{R}_\text{y}$ around the y-axis is calculated such that
\begin{equation}
    \mathbf{R}_\text{y} \cdot \mathbf{R}_\text{xz} \cdot \mathbf{w}^\text{c*}_o = (\eta, \zeta, 0) 
\end{equation}
with $\eta \geq 0$ for a specified keypoint index $o$. The total transformation between canonical and original frame is given by
\begin{equation}
    \mathbf{R}(\mathbf{w}^\text{rel}) = \mathbf{R}_\text{y} \cdot \mathbf{R}_\text{xz} \text{.}
\end{equation}
In order to deal appropriately with the symmetry between left and right hands, we flip right hands along the z-axis, which yields the side agnostic representation 
\begin{equation}
    \mathbf{w}^\text{c}_i =
   \left\{
    \begin{array}{ll}
        (x^\text{c*}_i, y^\text{c*}_i, z^\text{c*}_i)^\top  &\text{if its a left hand}\\
        (x^\text{c*}_i, y^\text{c*}_i, -z^\text{c*}_i)^\top &\text{if its a right hand}
    \end{array}
    \right.
\end{equation}
that resembles our proposed canonical coordinate system.
Given this canonical frame definition, we train our network to estimate the 3D coordinates within the canonical frame $\mathbf{w}^\text{c}$ and separately to estimate the rotation matrix $\mathbf{R}(\mathbf{w}^\text{rel})$, which we parameterize using axis-angle notation with three parameters.  Estimating the transformation $\mathbf{R}$ is equivalent to predicting the viewpoint of a given sample with respect to the canonical frame. Thus, we refer to the problem as \emph{viewpoint estimation}. 

The network architecture for the pose prior has two parallel processing streams; see Figure~\ref{fig:arch_poseprior}. The streams use an almost identical architecture given in the supplementary. They first process the stack of $J$ score maps in a series of $6$ convolutions with ReLU non-linearities. Information on whether the image shows a left or right hand is concatenated with the feature representation and processed further by two fully-connected layers. The streams end with a fully-connected layer with linear activation, which yields estimations for viewpoint $\mathbf{R}$ and canonical coordinates $\mathbf{w}^\text{c}$. Both estimations combined lead to an estimation of $\mathbf{w}^\text{rel}$.

%-------------------------------------------------------------------------
\subsection{Network training}
\label{subsec:network_training}

For training of \textit{HandSegNet} we apply standard softmax cross-entropy loss and $L_2$ loss for \textit{PoseNet}.
The \textit{PosePrior} network uses two loss terms. First a squared $L_2$ loss for the canonical coordinates
\begin{equation}
    \text{L}_{\text{c}} = \norm{\mathbf{w}^\text{c}_\text{gt} - \mathbf{w}^\text{c}_\text{pred}}^2
\end{equation}
based on the network predictions $\mathbf{w}^\text{c}_\text{pred}$ and the ground truth $\mathbf{w}^\text{c}_\text{gt}$. Secondly, a squared $L_2$ loss is imposed on the canonical transformation matrix:
\begin{equation}
    \text{L}_{\text{r}} = \norm{\mathbf{R}_\text{pred} - \mathbf{R}_\text{gt}}^2 \text{.}
\end{equation}
The total loss function is the unweighted sum of  $\text{L}_{\text{c}}$ and $\text{L}_{\text{r}}$.

We used Tensorflow \cite{abadi2016tensorflow} with the Adam solver \cite{adam_kingsma} for training. Details on the learning procedure are in the supplementary material.

%-------------------------------------------------------------------------
\section{Datasets for hand pose estimation}
\label{sec:dataset}
\subsection{Available datasets}
\label{subsec:available_datasets}

There are two available datasets that apply to our problem, as they provide RGB images and 3D pose annotation. 
The so-called {\em Stereo Hand Pose Tracking Benchmark} \cite{zhang_3d_2016} provides both 2D and 3D annotations of 21 keypoints for $18000$ stereo pairs with a resolution of $640\! \times \!480$. 
The dataset shows a single person's left hand in front of $6$ different backgrounds and under varying lighting conditions. 
We divided the dataset into an evaluation set of $3000$ images (\textbf{\textit{S-val}}) and a training set with $15000$ images (\textbf{\textit{S-train}}).

{\em Dexter} \cite{sridhar2016real} is a dataset providing $3129$ images showing two operators performing different kinds of manipulations with a cuboid in a restricted indoor setup. The dataset provides color images, depth maps, and annotations for fingertips and cuboid corners. The color images have a spatial resolution of $640\! \times \!320$. Due to the incomplete hand annotation, we use this dataset only for investigating the cross-dataset generalization of our network. We refer to this test set as \textbf{\textit{Dexter}}.

We downsampled both datasets to a resolution of $320\! \times \!240$ to be compatible with our rendered dataset. We transform our results back to coordinates in the original resolution, when we report pixel accuracies in the image domain.

The NYU Hand Pose Dataset by Tompson \etal~\cite{tompson_real-time_2014}, commonly used for hand pose estimation from depth images, does not apply to a color based approach, because only registered color images are provided. In the supplementary we show more evidence why this dataset cannot be used for our task.

\subsection{Rendered hand pose dataset}
\label{subsec:rendered_dataset}

The above datasets are not sufficient for training a deep network due to limited variation, number of available samples, and partially incomplete annotation. 
Therefore, we complement them with a new dataset for training. To avoid the known problem of poor labeling performance by human annotators in three-dimensional data, we utilize freely available 3D models of humans with corresponding animations from Mixamo \footnote{\href{http://www.mixamo.com}{http://www.mixamo.com}}. Then we used the open source software \textit{Blender} \footnote{\href{http://www.blender.org}{http://www.blender.org}} to render images. 
The dataset is publicly available online. 

\begin{figure}
\centering
\begin{subfigure}{.4\columnwidth}
  \centering
  \includegraphics[width=.9\textwidth]{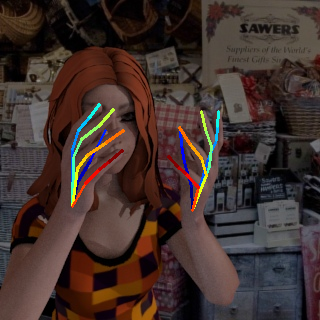}
\end{subfigure}%
\begin{subfigure}{.4\columnwidth}
  \centering
  \includegraphics[width=.9\textwidth]{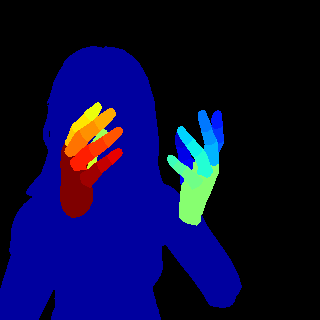}
\end{subfigure}
\caption{Our new dataset provides segmentation maps with $33$ classes: three for each finger, palm, person, and background. The 3D kinematic model of the hand provides $21$ keypoints per hand: 4 keypoints per finger and one keypoint close to the wrist.}
\label{fig:dataset_sample}
\end{figure}

Our dataset is built upon $20$ different characters performing $39$ actions. 
We split the data into a validation set (\textbf{\textit{R-val}}) and a training set (\textbf{\textit{R-train}}), where a character or action can exclusively be in one of the sets but not in the other. Our proposed split results into $16$ characters performing $31$ actions for training and $4$ characters with $8$ actions in the validation set. 

For each frame we randomly sample a new camera location, which is roughly located in a spherical vicinity around one of the hands. All hand centers lie approximately in a range between $40$cm and $65$cm from the camera center. Both left and right hands are equally likely and the camera is rotated to ensure that the hand is at least partially visible from the current viewpoint. After the camera location and orientation are fixed, we randomly sample one background image from a pool of $1231$ background images downloaded from Flickr \footnote{\href{http://www.flickr.com}{http://www.flickr.com}}. Those images show different kinds of scenes from cities and landscapes. We ensured that they do not contain persons. 

To maximize the visual diversity of the dataset, we randomize the following settings for each rendered frame: we apply lighting by $0$ to $2$ directional light sources and global illumination, such that the color of the sampled background image is roughly matched. Additionally we randomize light positions and intensities. Furthermore, we save our renderings using a lossy JPG compression with the quality factor being randomized from no compression up to $60\%$. We also randomized the effect of specular reflections on the skin.

In total our dataset provides $41258$ images for training and $2728$ images for evaluation with a resolution of $320\! \times \!320$ pixels. All samples come with full annotation of a $21$ keypoint skeleton model of each hand and additionally $33$ segmentation masks are available plus the background. As far as the segmentation masks are concerned there is a class for the human, one for each palm and each finger is composed by $3$ segments. Figure \ref{fig:dataset_sample} shows a sample from the dataset. Every finger is represented by 4 keypoints: the tip of the finger, two intermediate keypoints and the end located on the palm. Additionally, there is a keypoint located at the wrist of the model. For each of the hand keypoints, there is information if it is visible or occluded in the image. Also keypoint annotations in the camera pixel coordinate system and in camera centered world coordinates are given. The camera intrinsic matrix and a ground truth depth map are available, too, but were not used in this work.

%-------------------------------------------------------------------------
\section{Experiments}
\label{sec:experiments}

We evaluated all relevant parts of the overall approach: 
(1) the detection of hand keypoints of the \textit{PoseNet} with and without the 
hand segmentation network; (2) the 3D hand pose estimation and the learned 3D 
pose prior. Finally, we applied the hand pose estimation to a sign language 
recognition benchmark.

\subsection{Keypoint detection in 2D}
\label{subsec:keypoint_detection_2d}
\newcommand{\figureWidth}{.16}
\begin{figure*}
\centering
\begin{subfigure}{\figureWidth\textwidth}
  \centering
  \includegraphics[width=\textwidth]{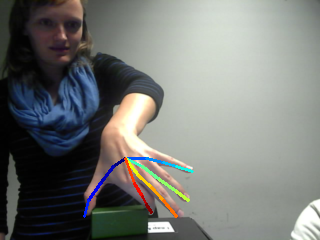}
\end{subfigure}
\begin{subfigure}{\figureWidth\textwidth}
  \centering
  \includegraphics[width=\textwidth]{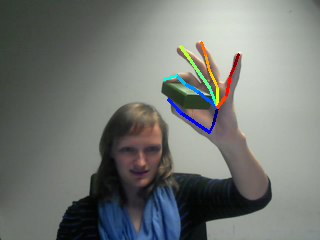}
\end{subfigure}
\begin{subfigure}{\figureWidth\textwidth}
  \centering
  \includegraphics[width=\textwidth]{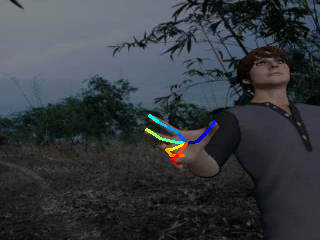}
\end{subfigure}
\begin{subfigure}{\figureWidth\textwidth}
  \centering
  \includegraphics[width=\textwidth]{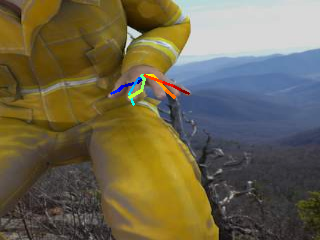}
\end{subfigure}
\begin{subfigure}{\figureWidth\textwidth}
  \centering
  \includegraphics[width=\textwidth]{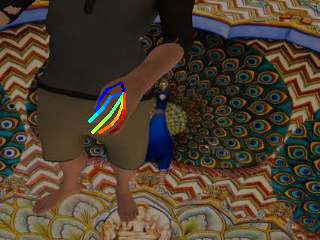}
\end{subfigure}
\begin{subfigure}{\figureWidth\textwidth}
  \centering
  \includegraphics[width=\textwidth]{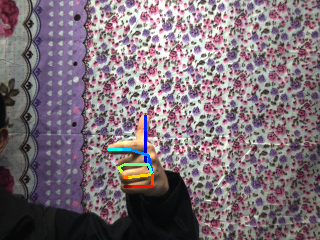}
\end{subfigure}

\begin{subfigure}{\figureWidth\textwidth}
  \centering
  \includegraphics[width=\textwidth]{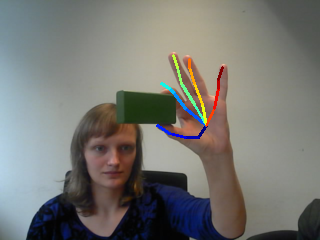}
\end{subfigure}
\begin{subfigure}{\figureWidth\textwidth}
  \centering
  \includegraphics[width=\textwidth]{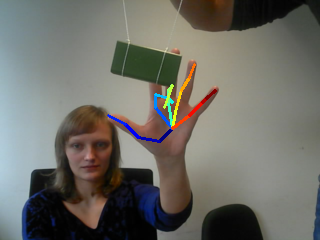}
\end{subfigure}
\begin{subfigure}{\figureWidth\textwidth}
  \centering
  \includegraphics[width=\textwidth]{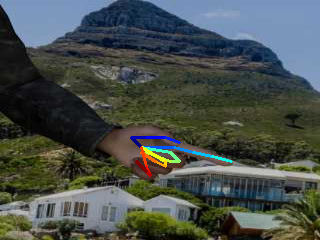}
\end{subfigure}
\begin{subfigure}{\figureWidth\textwidth}
  \centering
  \includegraphics[width=\textwidth]{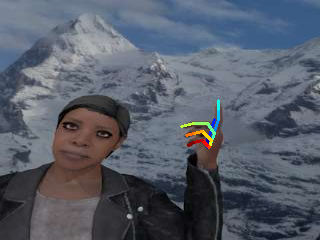}
\end{subfigure}
\begin{subfigure}{\figureWidth\textwidth}
  \centering
  \includegraphics[width=\textwidth]{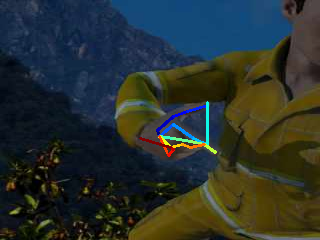}
\end{subfigure}
\begin{subfigure}{\figureWidth\textwidth}
  \centering
  \includegraphics[width=\textwidth]{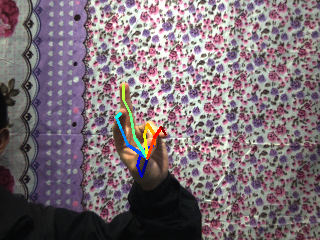}
\end{subfigure}
\caption{Exemplary 2D keypoint localization results. The first two columns show samples from \textit{Dexter}, the following three depict \textit{R-val} and the last one are samples from \textit{S-val}.}
\label{fig:results_2d}
\end{figure*}

\begin{figure}
\centering
    \includegraphics[width=0.45\textwidth, height=0.25\textwidth]{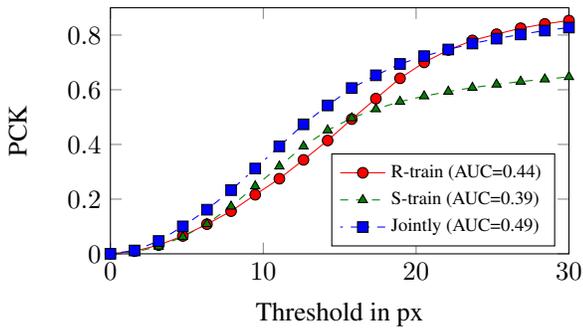}
\caption{Results on 2D keypoint estimation when using different training sets for \textit{PoseNet}. Shown is the percentage of correct keypoints (PCK) over a certain threshold in pixels evaluated on \textit{Dexter}. Jointly training on R-train and S-train yields the best results.}\label{fig:auc_curves_2d_dexter}
\end{figure}

Table \ref{tab:posenet2d_performance} shows the performance of \textit{PoseNet} on 2D keypoint estimation. We report the average endpoint error (EPE) in pixels and the area under the curve (AUC) on the percentage of correct keypoints (PCK) for different error thresholds; see Figure~\ref{fig:auc_curves_2d_dexter}.

\begin{table}
\begin{center}
\begin{tabular}{|c|c|c|c|c|}
\hline
\multicolumn{1}{|c|}{ } & \multicolumn{1}{c|}{ }  & \multicolumn{1}{c|}{AUC}  & \multicolumn{1}{c|}{EPE median}  & \multicolumn{1}{c|}{EPE mean} \\ \hline
\parbox[t]{2mm}{\multirow{2}{*}{\rotatebox[origin=c]{90}{GT}}} &
R-val & $0.724$ & $5.001$ & $9.135$ \\
& S-val & $0.817$ & $5.013$ & $5.522$ \\ \hline
\parbox[t]{2mm}{\multirow{3}{*}{\rotatebox[origin=c]{90}{Net}}} &
R-val & $0.635$ & $6.745$ & $18.741$ \\
& S-val & $0.762$ & $5.528$ & $18.581$ \\
& Dexter & $0.489$ & $13.684$ & $25.160$ \\ \hline
\end{tabular}
\caption{The top rows (GT) report performance for the \textit{PoseNet} operating on ground truth cropped hand images. The bottom rows (Net) show results when the hand crops are generated using \textit{HandSegNet}. \textit{PoseNet} was trained jointly on \textit{R-train} and \textit{S-train}, whereas \textit{HandSegNet} was only trained on \textit{R-train}. End point errors are reported in pixels with respect to the uncropped image and AUC is calculated over an error range from $0$ to $30$ pixels.\invisnote{exp-bwr55}} \label{tab:posenet2d_performance}
\end{center}
\end{table}

We evaluated two cases: one using images, where the hand is cropped with the ground truth oracle (GT), and one using the predictions from \textit{HandSegNet} for cropping (Net). The first case shows the performance of \textit{PoseNet} in isolation, while the second shows the performance of the complete 2D keypoint estimation. The difference between the median and the mean for the latter case show that \textit{HandSegNet} is reliable in most cases but is sometimes not able to segment the hand correctly, which makes the 2D keypoint prediction fail. 

The results show that the method works on our synthetic dataset (\textit{R-val}) and the stereo dataset (\textit{S-val}) equally well. The \textit{Dexter} dataset is more difficult because the dataset is different from the training set and because of frequent occlusions of the hand by the handled cube. We did not have samples with occlusion (apart from self-occlusion) in the training set. 

In Figure~\ref{fig:auc_curves_2d_dexter} we show that training on more diverse data helps cross-dataset generalization. While training only on our synthetic dataset \textit{R-train} yields much better results on \textit{Dexter} than training on the limited stereo dataset \textit{S-train}, training on \textit{R-train} and \textit{S-train} together yields the best results. Figure~\ref{fig:results_2d} shows some qualitative results of this configuration. Additional examples are in the supplementary.

\begin{figure*}
\centering
\begin{subfigure}{.15\textwidth}
  \centering
  \includegraphics[width=\textwidth]{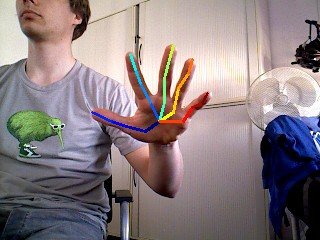}
\end{subfigure}
\begin{subfigure}{.15\textwidth}
  \centering
  \includegraphics[width=\textwidth]{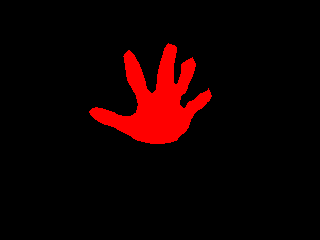}
\end{subfigure}
\begin{subfigure}{.15\textwidth}
  \centering
  \includegraphics[width=\textwidth]{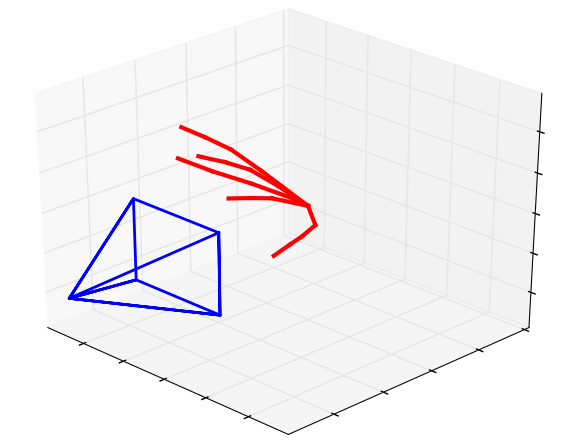}
\end{subfigure}
\begin{subfigure}{.15\textwidth}
  \centering
  \includegraphics[width=\textwidth]{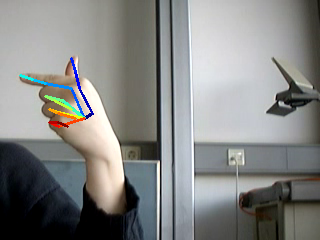}
\end{subfigure}
\begin{subfigure}{.15\textwidth}
  \centering
  \includegraphics[width=\textwidth]{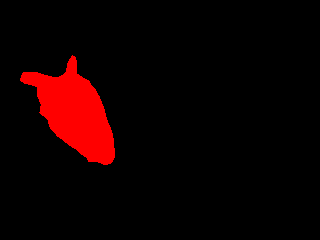}
\end{subfigure}
\begin{subfigure}{.15\textwidth}
  \centering
  \includegraphics[width=\textwidth]{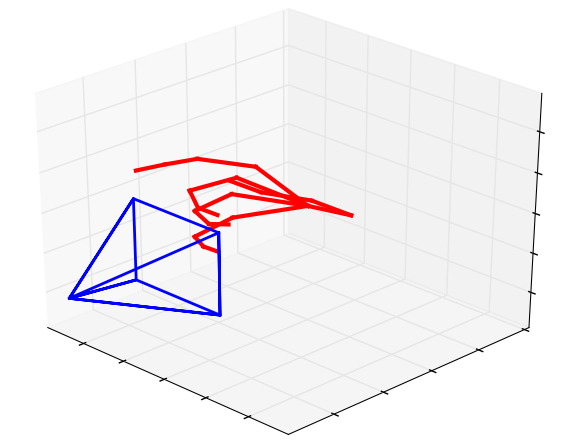}
\end{subfigure}

\begin{subfigure}{.15\textwidth}
  \centering
  \includegraphics[width=\textwidth]{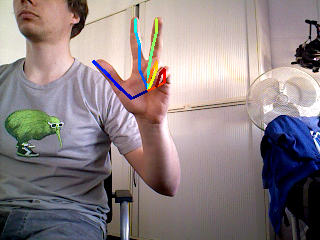}
\end{subfigure}
\begin{subfigure}{.15\textwidth}
  \centering
  \includegraphics[width=\textwidth]{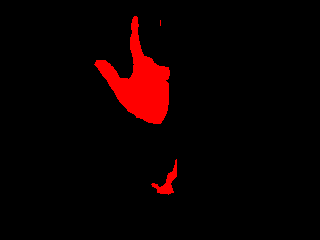}
\end{subfigure}
\begin{subfigure}{.15\textwidth}
  \centering
  \includegraphics[width=\textwidth]{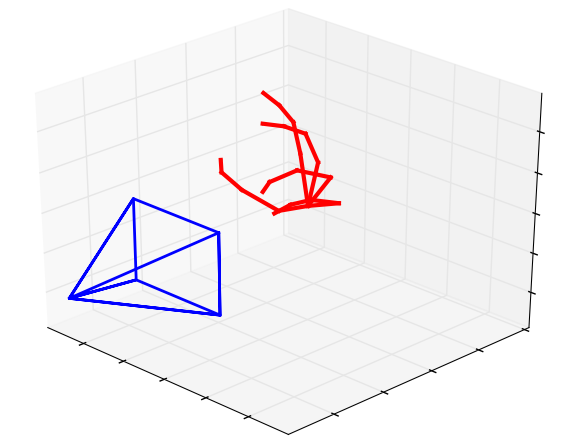}
\end{subfigure}
\begin{subfigure}{.15\textwidth}
  \centering
  \includegraphics[width=\textwidth]{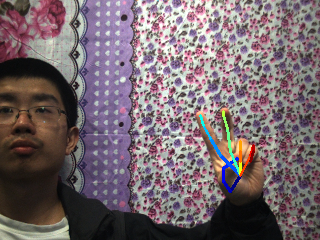}
\end{subfigure}
\begin{subfigure}{.15\textwidth}
  \centering
  \includegraphics[width=\textwidth]{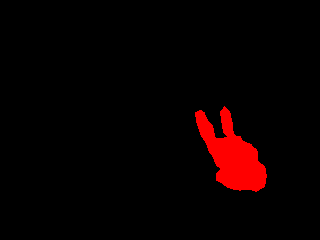}
\end{subfigure}
\begin{subfigure}{.15\textwidth}
  \centering
  \includegraphics[width=\textwidth]{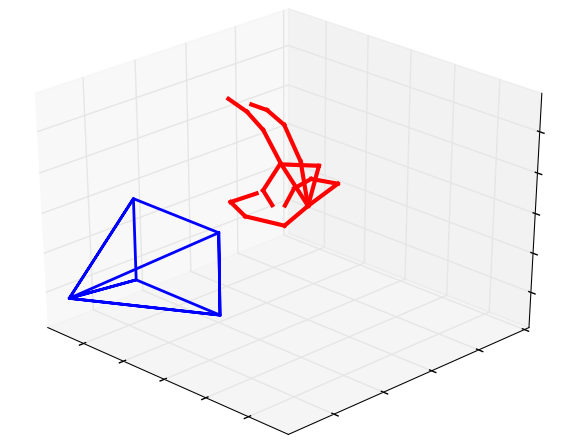}
\end{subfigure}
\caption{Qualitative examples of our complete system. Input to the network are color image and the information if its a left or right hand. The network estimates the hand segmentation mask, localizes keypoints in 2D and outputs the most likely 3D pose. The samples on the left hand side are from a dataset we recorded for qualitative evaluation, on the top right hand side is a sample from the sign language dataset and the bottom right sample is taken from \textit{S-val}. In the supplementary material we provide more qualitative examples. }
\label{fig:results3d}
\end{figure*}

\subsection{Lifting the estimation to 3D}
\label{subsec:lifting_2d_pose}

% Version in mm and percent subscript
\begin{table}
\begin{center}
\resizebox{\columnwidth}{!}{
  \begin{tabular}{|c|c|c|c|c|c|}
  \hline
  & Direct & Bottleneck & Local & NN  & \textbf{Prop.}\\
  \hline\hline
  R-train & $20.2_{\,9.2\%}$ & $21.1_{\,14\%}$ & $35.1_{\,90\%}$ & $0.0_{\,-100\%}$ & $18.5$ \\
  R-val & $20.9_{\,11.2\%}$ & $21.9_{\,16\%}$ & $39.1_{\,108\%}$ & $26.9_{\,43\%}$ & $18.8$ \\
  \hline
  \end{tabular}
}
\caption{Average median end point error per keypoint of the predicted 3D pose for different lifting approaches
given a noisy ground truth 2D pose. 
Networks were trained on \textit{R-train}. The results are reported in mm and the subscript gives the relative
performance to the proposed approach.\invisnote{exp r163, r164, r161}}\label{tab:lifting_gt_epe}

\end{center}
\end{table}

\subsubsection{Pose representation}
\newcommand{\figureWidthTwo}{.09}
\newcommand{\figureWidthThree}{.13}
\begin{figure}[ht]
\centering
%second row
\begin{subfigure}{\figureWidthTwo\textwidth}
  \centering
  \includegraphics[width=\textwidth]{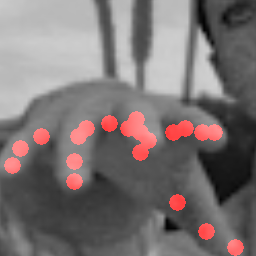}
\end{subfigure}
\begin{subfigure}{\figureWidthThree\textwidth}
  \centering
  \includegraphics[width=\textwidth]{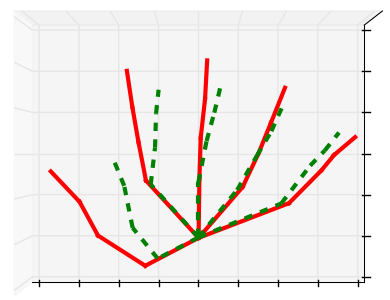}
\end{subfigure}
\begin{subfigure}{\figureWidthThree\textwidth}
  \centering
  \includegraphics[width=\textwidth]{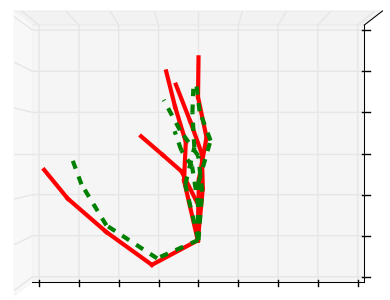}
\end{subfigure}

%third row
\begin{subfigure}{\figureWidthTwo\textwidth}
  \centering
  \includegraphics[width=\textwidth]{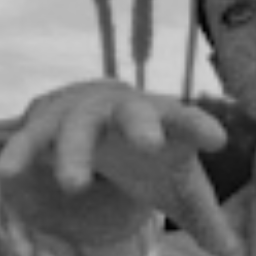}
\end{subfigure}
\begin{subfigure}{\figureWidthThree\textwidth}
  \centering
  \includegraphics[width=\textwidth]{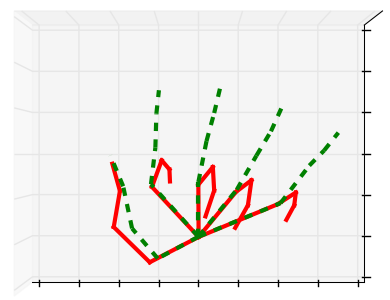}
\end{subfigure}
\begin{subfigure}{\figureWidthThree\textwidth}
  \centering
  \includegraphics[width=\textwidth]{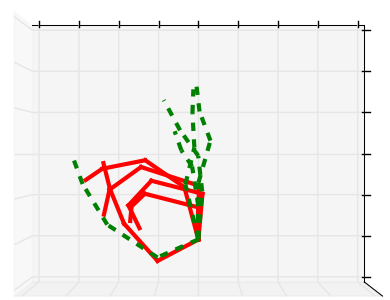}
\end{subfigure}

%forth row
\begin{subfigure}{\figureWidthTwo\textwidth}
  \centering
  \includegraphics[width=\textwidth]{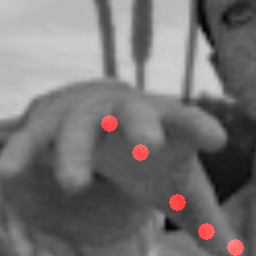}
\end{subfigure}
\begin{subfigure}{\figureWidthThree\textwidth}
  \centering
  \includegraphics[width=\textwidth]{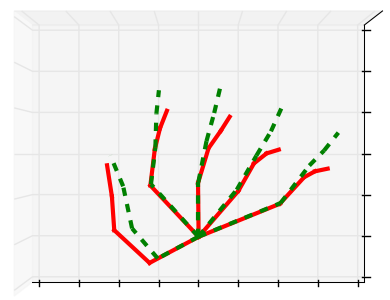}
\end{subfigure}
\begin{subfigure}{\figureWidthThree\textwidth}
  \centering
  \includegraphics[width=\textwidth]{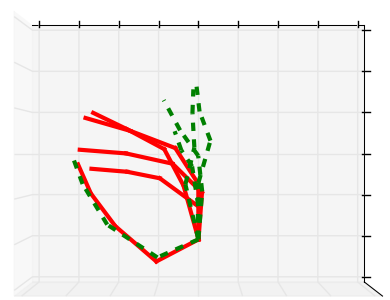}
\end{subfigure}

%fifth row
\begin{subfigure}{\figureWidthTwo\textwidth}
  \centering
  \includegraphics[width=\textwidth]{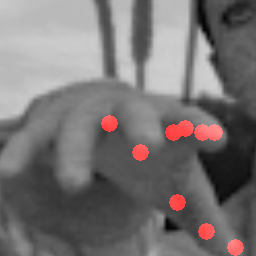}
\end{subfigure}
\begin{subfigure}{\figureWidthThree\textwidth}
  \centering
  \includegraphics[width=\textwidth]{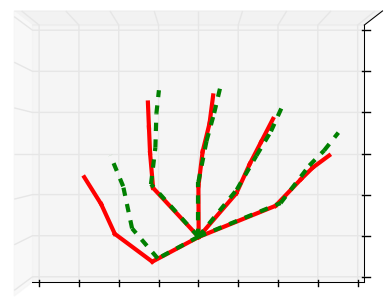}
\end{subfigure}
\begin{subfigure}{\figureWidthThree\textwidth}
  \centering
  \includegraphics[width=\textwidth]{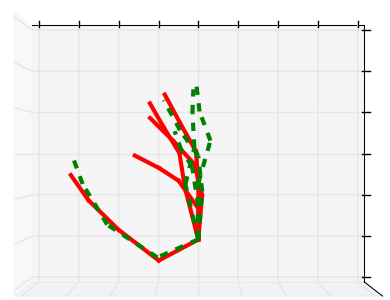}
\end{subfigure}
\caption{The left most column shows the input image as gray scale with the input score map overlayed as red dots. Every row corresponds to a separate forward pass of the network. The two columns to the right visualize the predicted 3D structure of the network from different viewpoints in canonical coordinates. Ground truth is displayed in dashed green and the network prediction is shown in solid red.}
\label{fig:sanity_of_pose_prior}
\end{figure}

We evaluated the proposed canonical frame representation for predicting the 3D hand pose from 2D keypoints by comparing it to several alternatives. All variants share a common base architecture that is identical to one stream of the \textit{PosePrior} proposed in \ref{subsubsec:pose_prior}.
They were trained on score maps $\mathbf{c}$ with a spatial resolution of $32$ by $32$ pixels. To avoid overfitting, we augmented the score maps by disturbing the keypoint location with Gaussian noise of variance $1.5$ pixel. Additionally the scoremaps are randomly scaled and translated. Table~\ref{tab:lifting_gt_epe} shows the resulting end point errors per keypoint.

The \textit{Direct} approach tries to lift the 2D keypoints directly to the full 3D coordinates $\mathbf{w}^\text{rel}$ without using a canonical frame. This is disadvantageous, because it is difficult for the network to learn separate the global rotation of the hand from the articulation.

The \textit{Bottleneck} approach is inspired by Oberweger \etal~\cite{oberweger_hands_2015}, who introduced a bottleneck layer before estimating the coordinates. We inserted an additional FC layer before the final FC output layer, parameterize it as in Oberweger \etal with $30$ channels and linear activation. The outcome was not better than with the \textit{Direct} approach. 
\vfill\null
The \textit{Local} approach incorporates the kinematic model of the hand and uses the network to estimate articulation parameters of the model. We generalize \cite{zhou_model-based_2016} by estimating not only the angles but also the bone length. The network is trained to estimate two angles and one length per keypoint, which results in $63$ parameters. The angles express rotations in a bone local coordinate system. This approach only works if the hand is always shown from the same direction, but cannot capture the global pose of the hand. 

Finally, the \textit{NN} approach matches the 2D keypoints to the most similar sample from the training set and retrieves the 3D coordinates from this sample \cite{chen_3d_2016}. While this approach trivially works best on the training set, it does not generalize well to new samples. 

The generalization of the other approaches is quite good showing similar errors for both the training and the validation set. The proposed approach from \ref{subsubsec:pose_prior} worked best and was used for the following experiments.

\subsubsection{Analysis of the learned prior}
\label{subsubsec:pose_prior_exp}

\begin{figure}
\centering
    \includegraphics[width=0.45\textwidth, height=0.3\textwidth]{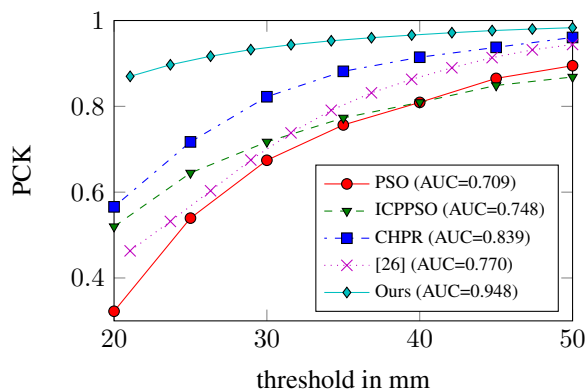}
\caption{Results for our complete system on \textit{S-val} compared to approaches from \cite{zhang_3d_2016} and \cite{zhao_simple_2016}. Shown is the percentage of correct keypoints (PCK) over respective thresholds in mm. \textit{PoseNet} and \textit{PosePrior} are trained on \textit{S-train} and \textit{R-train}, whereas the \textit{HandSegNet} is trained on \textit{R-train}.}\label{fig:auc_stereo}
\end{figure}

To examine the 3D prior learned by the network we input score maps that lack keypoints and Figure~\ref{fig:sanity_of_pose_prior} shows the 3D pose prediction from two different viewpoints. The extreme case, with no keypoints provided as input at all, shows the canonical prior learned by the network. As more keypoints are added, the network adjusts the predicted pose to this additional evidence. This experiment also simulates the situation of occluded 2D keypoints and demonstrates that the learned prior allows the network to still retrieve reasonable poses.

\subsubsection{Comparison to literature}

Since there is no work on 3D hand pose estimation from RGB images yet, we cannot compare to alternative approaches.
To still relate our results coarsely to literature, we compare them to 
Zhang \etal~\cite{zhang_3d_2016}, who provide results in mm for state-of-the-art 3D hand pose tracking on depth data. They run their experiments on the stereo dataset \textit{S-val}, which also contains RGB images.
Since in contrast to Zhang \etal our approach does not use the depth data, it still comes with ambiguities with regard to scale and absolute depth. 
Thus, we accessed the absolute position of the root keypoint and the scale of the hand to shift and scale our predicted 3D hand pose, which yields metric world coordinates $\mathbf{w}$ by using \eqref{eq:scale_eq} and \eqref{eq:rel_eq}.  
For this experiment we trained \textit{PosePrior} on score maps predicted by \textit{PoseNet} using the same schedule as for the experiment in section \ref{subsubsec:pose_prior_exp}. \textit{PoseNet} is trained separately as described in \ref{subsec:keypoint_detection_2d} and then kept fixed.
Figure \ref{fig:auc_stereo} shows that our approach largely outperforms the approaches presented in Zhang \etal~\cite{zhang_3d_2016} although we use the depth map only for rescaling and shifting in the end. 
Additionally we report results of the lifting approach presented by Zhao \etal~\cite{zhao_simple_2016} in conjunction with our \textit{PoseNet}, which we train in a similar manner.
Results are inferior to the proposed \textit{PosePrior}. We believe the reason is that using score maps as input for the lifting is advantageous over coordinates,
because it can handle ambiguities in hand keypoint detection.
Qualitative 3D examples on three different datasets with the complete processing pipeline are shown in Figure \ref{fig:results3d}.

\subsection{Sign language recognition}
\label{subsec:sign_language_recognition}
Previous hand pose estimation approaches depending on depth data cannot be applied to most sign language recognition datasets, as they only come with color images. As a last exemplary experiment, we used our hand pose estimation system and trained a classifier for gesture recognition on top of it. The classifier is a fully connected three layer network with ReLU activation functions; c.f. the supplemental material for the network details. 

We report results on the so-called {\em RWTH German Fingerspelling Database} \cite{dreuw06smvp}. It contains $35$ gestures representing the letters of the alphabet, German umlauts, and the numbers from one to five. The dataset comprises $20$ different persons, who did two recordings each for every gesture. Most of the gestures are static except for the ones for the letters J, Z, Ä, Ö, and Ü, which are dynamic. In order to keep this experiment simple, we ran the experiments on the subset restricted to $30$ static gestures.

The database contains recordings by two different cameras, but we used only one camera. 
The short video sequences have a resolution of $320\! \times \!240$ pixels.
We used the middle frame from each video sequence as color image and the gesture class labels as training data.
This dataset has $1160$ images, which we separated by signers into a validation set with $232$ images and a training set with $928$ images. We resized image to $320\! \times \!320$ pixels and trained on randomly sampled $256\! \times \!256$ crops. Because the images were taken from a compressed video stream they exhibit significant compression artifacts previously unseen by our networks. Thus, we labeled $50$ images from the training set with hand keypoints, which we used to fine-tune our \textit{PoseNet} upfront. Afterwards the pose estimation part is kept fixed and we solely train the \textit{GestureNet}. Table \ref{tab:slr_results} shows that our system archives comparable results to Dreuw \etal~\cite{dreuw06smvp} on the subset of gestures we used for the comparison.

%-------------------------------------------------------------------------
\section{Conclusions}
\label{sec:conclusion}
We have presented the first learning based system to estimate 3D hand pose from a single image. 
We contributed a large synthetic dataset that enabled us to train a network successfully on the task. We have shown that the network learned a 3D pose prior that allows it to predict reasonable 3D hand poses from 2D keypoints in real world images.
While the performance of the network is even competitive to approaches that use depth maps, there is still much room for improvements. The performance seems mostly limited by the lack of an annotated large scale dataset with real-world images and diverse pose statistics. 

\begin{table}
\begin{center}
\begin{tabular}{|l|c|}
\hline
Method & Word error rate \\
\hline\hline
Dreuw \etal~\cite{dreuw06smvp} & $35.7$ \%\\
\hline
Dreuw on subset \cite{rwth_fingerspelling_db} & $36.56$ \%\\
Ours 3D & $33.2$ \%\\
\hline
\end{tabular}
\end{center}
\caption{Word error rates in percent on the RWTH German Fingerspelling Database 
subset of non dynamic gestures. Results for Dreuw \etal~\cite{dreuw06smvp}  on 
the subset from \cite{rwth_fingerspelling_db}.}
\label{tab:slr_results}
\end{table}

\section*{Acknowledgements}
We gratefully acknowledge funding by the 
Baden-Württemberg Stiftung as part of the projects ROTAH and RatTrack. 
Also we thank Nikolaus Mayer, Benjamin Ummenhofer and Maxim Tatarchenko
for valuable ideas and many fruitful discussions.

\clearpage
{\small
\bibliographystyle{ieee}
\bibliography{papers}
}

\clearpage
\appendix
% \section*{Supplementary}
% \label{sec:supplementary}

\title{Supplementary Material:\\ Learning to Estimate 3D Hand Pose from Single RGB Images}

\author{Christian Zimmermann, Thomas Brox\\
University of Freiburg\\
{\tt\small \{zimmermann, brox\}@cs.uni-freiburg.de}
}
\maketitle

% Input supplementary

\section{HandSegNet architecture and learning schedule}

Table \ref{tab:handsegnet_architecture} contains the architecture used for \textit{HandSegNet}. 
It was trained for hand segmentation on \textit{R-train} with a batch size of 8 and using ADAM solver \cite{adam_kingsma}. 
The network was initialized using weights of Wei \etal~\cite{wei_convolutional_2016} for layers 1 to 16 and then trained for $40000$ iterations 
using a standard softmax cross-entropy loss. The learning rate was $1\cdot10^{-5}$ for the first $20000$ iterations, $1\cdot10^{-6}$ for following $10000$ iterations 
and $1\cdot10^{-7}$ until the end. Except for random color hue augmentation of $0.1$ no data augmentation was used. From the $320\!\times\!320$ pixel images of the training
set a $256\!\times\!256$ crop was taken randomly.

\begin{table}[!h]
\begin{center}
\begin{tabular}{|c|c|c|c|}
\hline
id &Name & Kernel & Dimensionality \\
\hline\hline
& Input image & - & $256\!\times\!256\!\times\!3$\\
\hline
1 & Conv. + ReLU & $3\!\times\!3$ & $256\!\times\!256\!\times\!64$ \\
2 & Conv. + ReLU & $3\!\times\!3$ & $256\!\times\!256\!\times\!64$ \\
3 & Maxpool & $4\!\times\!4$ & $128\!\times\!128\!\times\!64$ \\
4 & Conv. + ReLU & $3\!\times\!3$ & $128\!\times\!128\!\times\!128$ \\
5 & Conv. + ReLU & $3\!\times\!3$ & $128\!\times\!128\!\times\!128$ \\
6 & Maxpool & $4\!\times\!4$ & $64\!\times\!64\!\times\!128$ \\
7 & Conv. + ReLU & $3\!\times\!3$ & $64\!\times\!64\!\times\!256$ \\
8 & Conv. + ReLU & $3\!\times\!3$ & $64\!\times\!64\!\times\!256$ \\
9 & Conv. + ReLU & $3\!\times\!3$ & $64\!\times\!64\!\times\!256$ \\
10 & Conv. + ReLU & $3\!\times\!3$ & $64\!\times\!64\!\times\!256$ \\
11 & Maxpool & $4\!\times\!4$ & $32\!\times\!32\!\times\!256$ \\
12 & Conv. + ReLU & $3\!\times\!3$ & $32\!\times\!32\!\times\!512$ \\
13 & Conv. + ReLU & $3\!\times\!3$ & $32\!\times\!32\!\times\!512$ \\
14 & Conv. + ReLU & $3\!\times\!3$ & $32\!\times\!32\!\times\!512$ \\
15 & Conv. + ReLU & $3\!\times\!3$ & $32\!\times\!32\!\times\!512$ \\
16 & Conv. + ReLU & $3\!\times\!3$ & $32\!\times\!32\!\times\!512$ \\
17 & Conv. & $1\!\times\!1$ & $32\!\times\!32\!\times\!2$ \\
18 & Bilinear Upsampling & - & $256\!\times\!256\!\times\!2$ \\
19 & Argmax & - & $256\!\times\!256\!\times\!1$ \\
\hline
& Hand mask & - & $256\!\times\!256\!\times\!1$\\
\hline
\end{tabular}
\caption{Network architecture of the proposed \textit{HandSegNet} network. Except for input and hand mask output every row of the table gives a data tensor of the network and the operations that produced it.\label{tab:handsegnet_architecture}} 
\end{center}
\end{table}

\section{PoseNet architecture and learning schedule}

Table \ref{tab:posenet_architecture} contains the architecture used for \textit{PoseNet}. In all cases it was trained with a batch size of 8 and using ADAM solver \cite{adam_kingsma}. The initial 16 layers of the network are initialized using weights of Wei \etal~\cite{wei_convolutional_2016} all others are randomly initialized \invisnote{with Xavier}. The network is trained for $30000$ iterations using a $L_2$ loss. The learning rate is $1\cdot 10^{-4}$ for the first $10000$ iterations, $1\cdot 10^{-5}$ for following $10000$ iterations and $1\cdot 10^{-6}$ until the end.
For ground truth generation of the score maps we use normal distributions with a variance of $25$ pixels and the mean being equal to the given keypoint location. We normalize the resulting maps such that each map contains values from $0$ to $1$, if there is a keypoint visible. For invisible keypoints the map is zero everywhere.

We train \textit{PoseNet} on axis aligned crops that are resized to a resolution of $256\!\times\!256$ pixels by bilinear interpolation. The bounding box is chosen such that all keypoints of a single hand are contained within the crop. We augment the cropping procedure by modifying the calculated bounding box in two ways. First, we add noise to the calculated center of the bounding box, which is sampled from a zero mean normal distribution with variance of $10$ pixels. The size of the bounding box is changed accordingly to still contain all hand keypoints. Second we find it helpful to improve generalization performance by adding a bit of noise on the coordinates used to generate the score maps. Therefore, we add a normal distribution of zero mean and variance $1.5$ to the ground truth keypoint coordinates, whereas each keypoint is sampled independently. Additionally we apply random contrast augmentation with a scaling factor between $0.5$ and $1.0$, which is sampled from a uniform distribution.

\begin{table}[!h]
\begin{center}
\begin{tabular}{|c|c|c|c|}
\hline
id &Name & Kernel & Dimensionality \\
\hline\hline
& Input image & - & $256\!\times\!256\!\times\!3$\\
\hline
1 & Conv. + ReLU & $3\!\times\!3$ & $256\!\times\!256\!\times\!64$ \\
2 & Conv. + ReLU & $3\!\times\!3$ & $256\!\times\!256\!\times\!64$ \\
3 & Maxpool & $4\!\times\!4$ & $128\!\times\!128\!\times\!64$ \\
4 & Conv. + ReLU & $3\!\times\!3$ & $128\!\times\!128\!\times\!128$ \\
5 & Conv. + ReLU & $3\!\times\!3$ & $128\!\times\!128\!\times\!128$ \\
6 & Maxpool & $4\!\times\!4$ & $64\!\times\!64\!\times\!128$ \\
7 & Conv. + ReLU & $3\!\times\!3$ & $64\!\times\!64\!\times\!256$ \\
8 & Conv. + ReLU & $3\!\times\!3$ & $64\!\times\!64\!\times\!256$ \\
9 & Conv. + ReLU & $3\!\times\!3$ & $64\!\times\!64\!\times\!256$ \\
10 & Conv. + ReLU & $3\!\times\!3$ & $64\!\times\!64\!\times\!256$ \\
11 & Maxpool & $4\!\times\!4$ & $32\!\times\!32\!\times\!256$ \\
12 & Conv. + ReLU & $3\!\times\!3$ & $32\!\times\!32\!\times\!512$ \\
13 & Conv. + ReLU & $3\!\times\!3$ & $32\!\times\!32\!\times\!512$ \\
14 & Conv. + ReLU & $3\!\times\!3$ & $32\!\times\!32\!\times\!512$ \\
15 & Conv. + ReLU & $3\!\times\!3$ & $32\!\times\!32\!\times\!512$ \\
16 & Conv. + ReLU & $3\!\times\!3$ & $32\!\times\!32\!\times\!512$ \\
17 & Conv. & $1\!\times\!1$ & $32\!\times\!32\!\times\!21$ \\
\hline
18 & Concat(16, 17) & - & $32\!\times\!32\!\times\!533$\\
19 & Conv. + ReLU & $7\!\times\!7$ & $32\!\times\!32\!\times\!128$ \\
20 & Conv. + ReLU & $7\!\times\!7$ & $32\!\times\!32\!\times\!128$ \\
21 & Conv. + ReLU & $7\!\times\!7$ & $32\!\times\!32\!\times\!128$ \\
22 & Conv. + ReLU & $7\!\times\!7$ & $32\!\times\!32\!\times\!128$ \\
23 & Conv. + ReLU & $7\!\times\!7$ & $32\!\times\!32\!\times\!128$ \\
24 & Conv. & $1\!\times\!1$ & $32\!\times\!32\!\times\!21$ \\
\hline
25 & Concat(16, 17, 24) & - & $32\!\times\!32\!\times\!554$\\
26 & Conv. + ReLU & $7\!\times\!7$ & $32\!\times\!32\!\times\!128$ \\
27 & Conv. + ReLU & $7\!\times\!7$ & $32\!\times\!32\!\times\!128$ \\
28 & Conv. + ReLU & $7\!\times\!7$ & $32\!\times\!32\!\times\!128$ \\
29 & Conv. + ReLU & $7\!\times\!7$ & $32\!\times\!32\!\times\!128$ \\
30 & Conv. + ReLU & $7\!\times\!7$ & $32\!\times\!32\!\times\!128$ \\
31 & Conv. & $1\!\times\!1$ & $32\!\times\!32\!\times\!21$ \\
\hline
\end{tabular}
\caption{Network architecture of the \textit{PoseNet} network. Except for input every row of the table represents a data tensor of the network and the operations that produced it. Outputs of the network are are predicted score maps $\mathbf{c}$ from layers 17, 24 and 31.\label{tab:posenet_architecture}} 
\end{center}
\end{table}

\section{PosePrior architecture}

Table \ref{tab:poseprior_architecture} contains the architecture used for each stream of the \textit{PosePrior}. It uses $6$ convolutional layers followed by two fully-connected layers. All use ReLU activation function and the fully-connected layers have a dropout probability of $0.2$ to randomly drop a neuron. Preceeding to the first FC layer, information about the hand side is concatenated to the flattened feature representation calculated by the convolutional layers. All drops in spatial dimension are due to strided convolutions. The network ceases with a fully-connected layer that estimates $P$ parameters, where $P=3$ for Viewpoint estimation and $P=63$ for the coordinate estimation stream.

\begin{table}[!h]
\begin{center}
\begin{tabular}{|c|c|c|c|}
\hline
id & Name & Kernel & Dimensionality \\
\hline\hline
& Input & - & $32\!\times\!32\!\times\!21$\\
\hline
1 & Conv. + ReLU & $3\!\times\!3$ & $32\!\times\!32\!\times\!32$ \\
2 & Conv. + ReLU & $3\!\times\!3$ & $16\!\times\!16\!\times\!32$ \\
3 & Conv. + ReLU & $3\!\times\!3$ & $16\!\times\!16\!\times\!64$ \\
4 & Conv. + ReLU & $3\!\times\!3$ & $8\!\times\!8\!\times\!64$ \\
5 & Conv. + ReLU & $3\!\times\!3$ & $8\!\times\!8\!\times\!128$ \\
6 & Conv. + ReLU & $3\!\times\!3$ & $4\!\times\!4\!\times\!128$ \\
7 & Reshape + Concat & - & $130$ \\
8 & FC + ReLU + Drop(0.2) & - & $512$ \\
9 & FC + ReLU + Drop(0.2) & - & $512$ \\
10 & FC & - & $512$ \\
\hline
& Output & - & $P$ \\
\hline
\end{tabular}
\caption{Network architecture of a single stream for the proposed \textit{PosePrior} network. Except for input and output every row of the table gives a data tensor of the network and the operations that produced it. Reduction in the spatial dimension is due to stride in the convolutions. $P$ is the number of estimated parameters and is $P=3$ for Viewpoint estimation and $P=63$ for the coordinate estimation stream.} \label{tab:poseprior_architecture}
\end{center}
\end{table}

\section{GestureNet architecture and learning schedule}
We train the \textit{GestureNet} using Adam solver, a batch size of $8$ and an initial learning rate of $1\cdot 10^{-4}$ which drops by one decade at $15000$ and $20000$ iterations. Training is finished at iteration $30000$. The network is trained with a standard softmax cross-entropy loss on randomly cropped $256\!\times\!256$ images.

\begin{table}
\begin{center}
\begin{tabular}{|c|c|c|}
\hline
id &Name & Dimensionality \\
\hline\hline
& Input $\mathbf{c}^\text{rel}$ & $63$\\
\hline
1 & FC + ReLU + Dropout(0.2) & $512$ \\
2 & FC + ReLU + Dropout(0.2) & $512$ \\
3 & FC & $35$ \\
\hline
\end{tabular}
\caption{Network architecture of the \textit{GestureNet} used for our experiments. All layers were initialized randomly. Probability to drop a neuron in the indicated layers is set to $0.2$.} \label{tab:gesturenet_architecture}
\end{center}
\end{table}

\begin{figure*}[!h]
\centering
\begin{subfigure}{.13\textwidth}
  \centering
  \includegraphics[width=\textwidth]{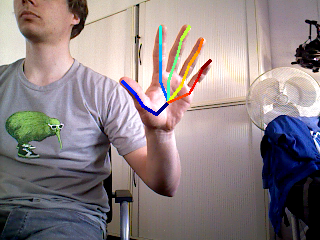}
\end{subfigure}
\begin{subfigure}{.13\textwidth}
  \centering
  \includegraphics[width=\textwidth]{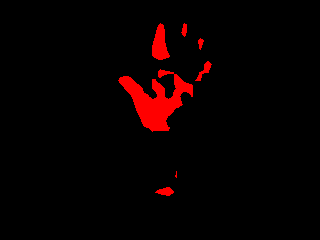}
\end{subfigure}
\begin{subfigure}{.13\textwidth}
  \centering
  \includegraphics[width=\textwidth]{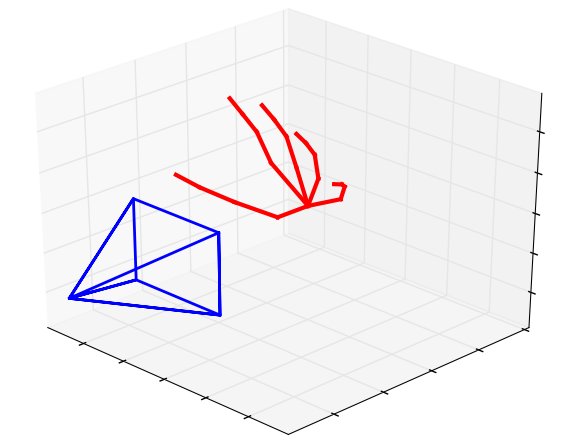}
\end{subfigure}
\begin{subfigure}{.13\textwidth}
  \centering
  \includegraphics[width=\textwidth]{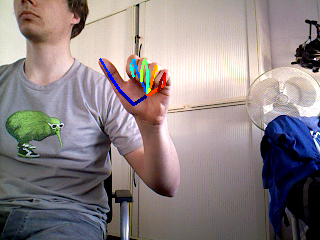}
\end{subfigure}
\begin{subfigure}{.13\textwidth}
  \centering
  \includegraphics[width=\textwidth]{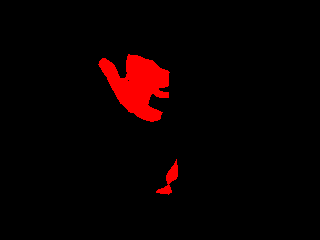}
\end{subfigure}
\begin{subfigure}{.13\textwidth}
  \centering
  \includegraphics[width=\textwidth]{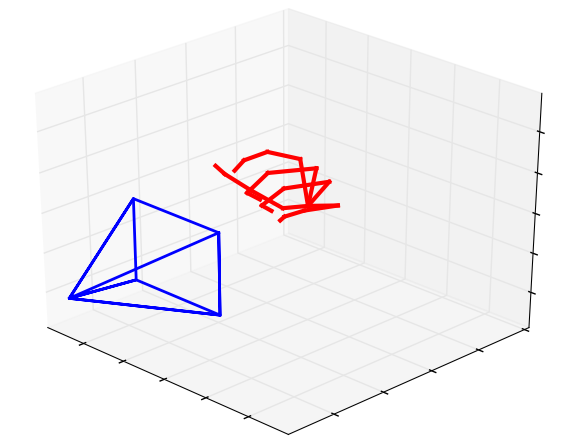}
\end{subfigure}

\begin{subfigure}{.13\textwidth}
  \centering
  \includegraphics[width=\textwidth]{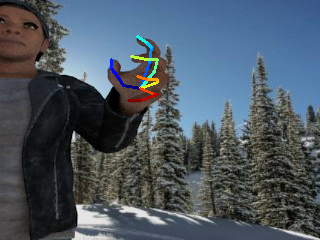}
\end{subfigure}
\begin{subfigure}{.13\textwidth}
  \centering
  \includegraphics[width=\textwidth]{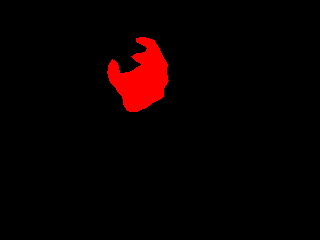}
\end{subfigure}
\begin{subfigure}{.13\textwidth}
  \centering
  \includegraphics[width=\textwidth]{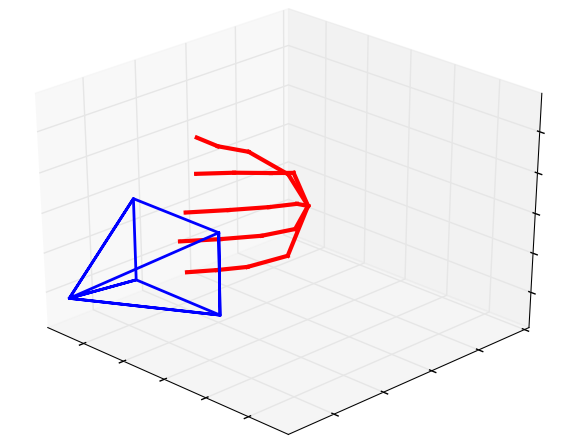}
\end{subfigure}
\begin{subfigure}{.13\textwidth}
  \centering
  \includegraphics[width=\textwidth]{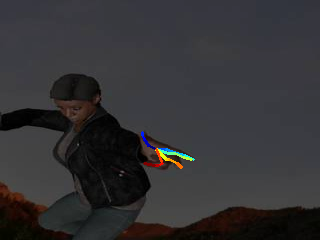}
\end{subfigure}
\begin{subfigure}{.13\textwidth}
  \centering
  \includegraphics[width=\textwidth]{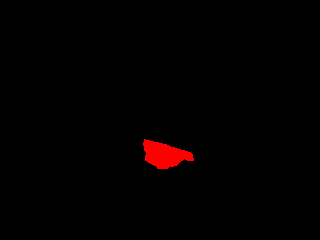}
\end{subfigure}
\begin{subfigure}{.13\textwidth}
  \centering
  \includegraphics[width=\textwidth]{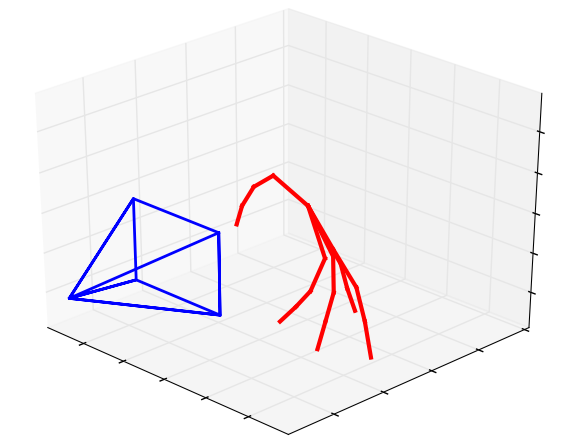}
\end{subfigure}

\begin{subfigure}{.13\textwidth}
  \centering
  \includegraphics[width=\textwidth]{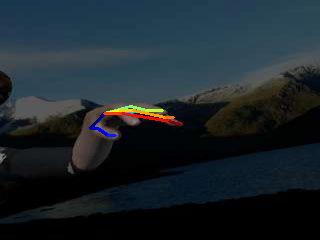}
\end{subfigure}
\begin{subfigure}{.13\textwidth}
  \centering
  \includegraphics[width=\textwidth]{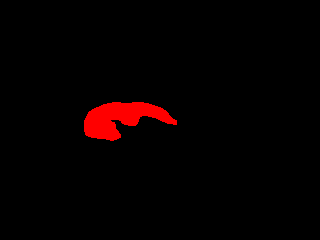}
\end{subfigure}
\begin{subfigure}{.13\textwidth}
  \centering
  \includegraphics[width=\textwidth]{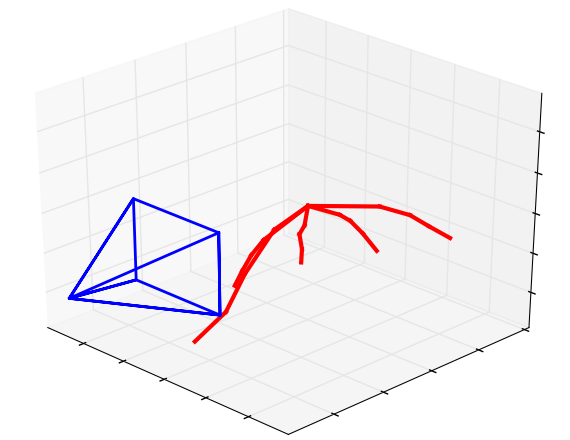}
\end{subfigure}
\begin{subfigure}{.13\textwidth}
  \centering
  \includegraphics[width=\textwidth]{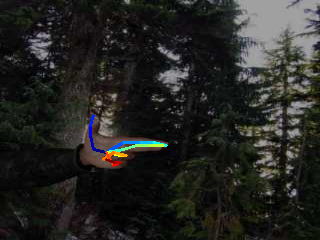}
\end{subfigure}
\begin{subfigure}{.13\textwidth}
  \centering
  \includegraphics[width=\textwidth]{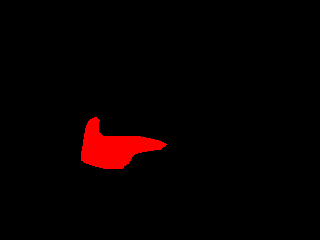}
\end{subfigure}
\begin{subfigure}{.13\textwidth}
  \centering
  \includegraphics[width=\textwidth]{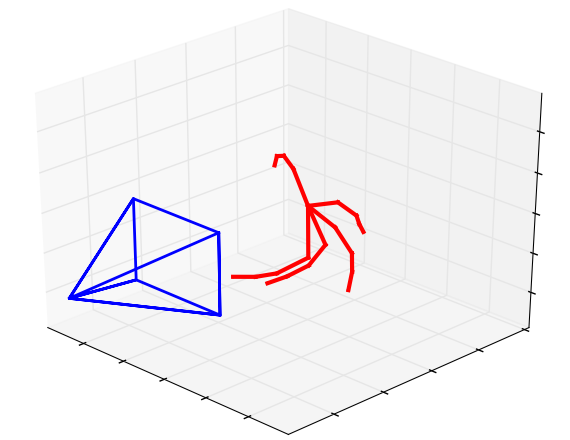}
\end{subfigure}

\begin{subfigure}{.13\textwidth}
  \centering
  \includegraphics[width=\textwidth]{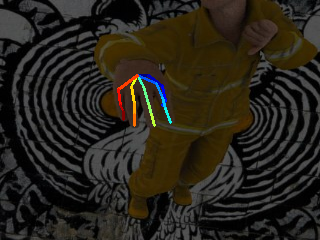}
\end{subfigure}
\begin{subfigure}{.13\textwidth}
  \centering
  \includegraphics[width=\textwidth]{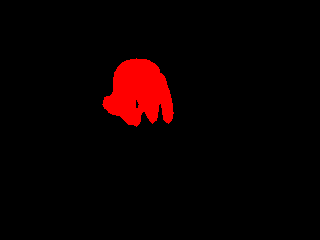}
\end{subfigure}
\begin{subfigure}{.13\textwidth}
  \centering
  \includegraphics[width=\textwidth]{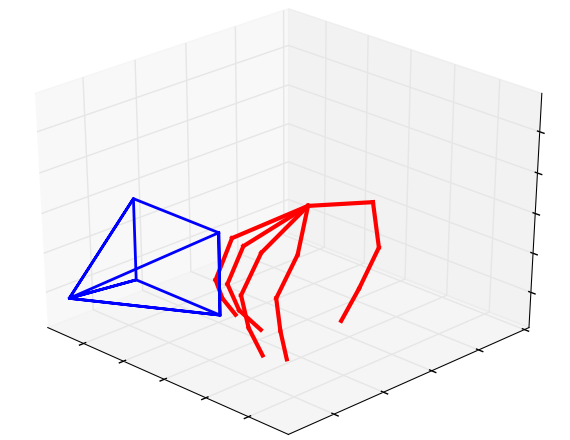}
\end{subfigure}
\begin{subfigure}{.13\textwidth}
  \centering
  \includegraphics[width=\textwidth]{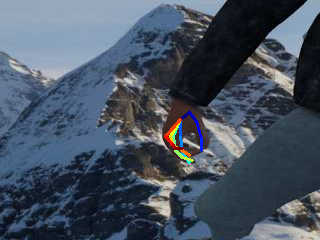}
\end{subfigure}
\begin{subfigure}{.13\textwidth}
  \centering
  \includegraphics[width=\textwidth]{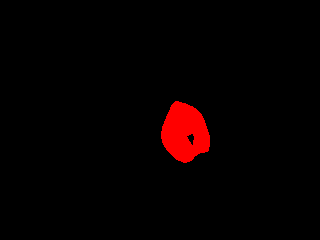}
\end{subfigure}
\begin{subfigure}{.13\textwidth}
  \centering
  \includegraphics[width=\textwidth]{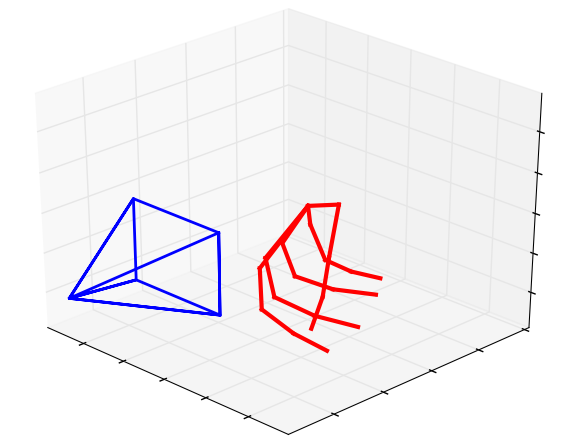}
\end{subfigure}

\begin{subfigure}{.13\textwidth}
  \centering
  \includegraphics[width=\textwidth]{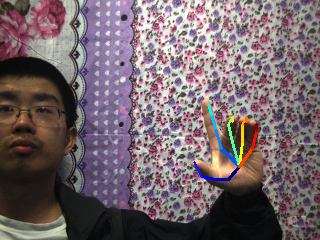}
\end{subfigure}
\begin{subfigure}{.13\textwidth}
  \centering
  \includegraphics[width=\textwidth]{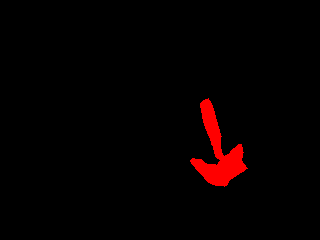}
\end{subfigure}
\begin{subfigure}{.13\textwidth}
  \centering
  \includegraphics[width=\textwidth]{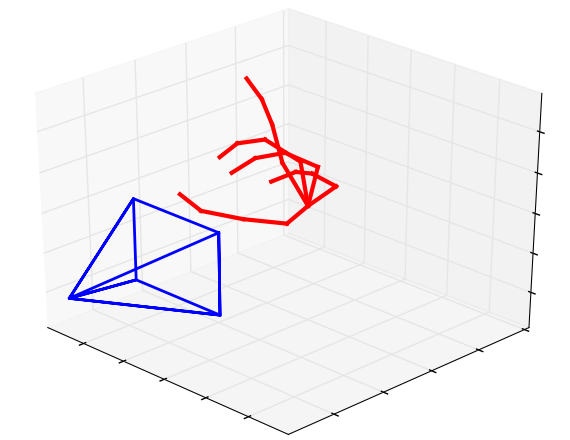}
\end{subfigure}
\begin{subfigure}{.13\textwidth}
  \centering
  \includegraphics[width=\textwidth]{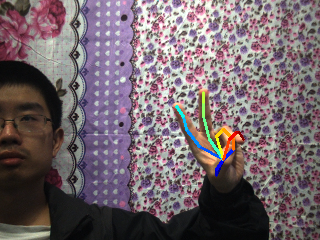}
\end{subfigure}
\begin{subfigure}{.13\textwidth}
  \centering
  \includegraphics[width=\textwidth]{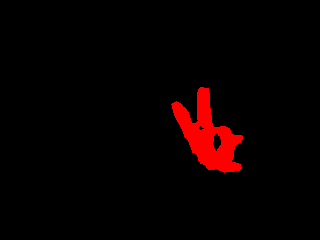}
\end{subfigure}
\begin{subfigure}{.13\textwidth}
  \centering
  \includegraphics[width=\textwidth]{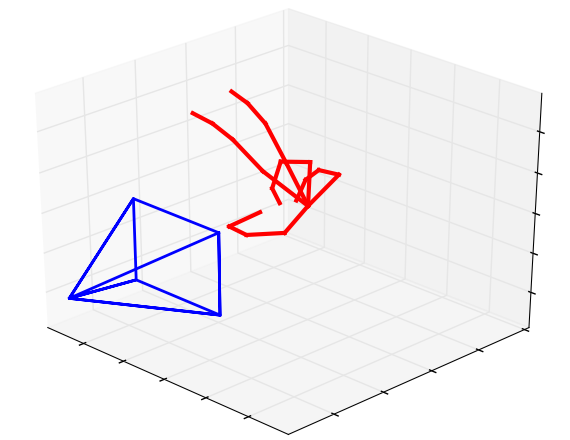}
\end{subfigure}

\begin{subfigure}{.13\textwidth}
  \centering
  \includegraphics[width=\textwidth]{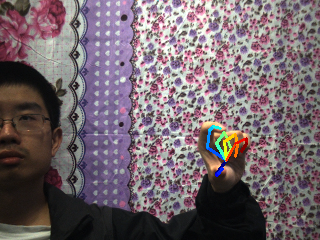}
\end{subfigure}
\begin{subfigure}{.13\textwidth}
  \centering
  \includegraphics[width=\textwidth]{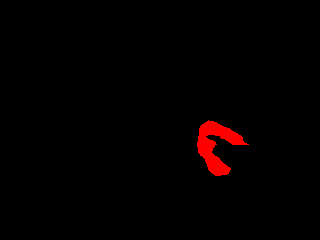}
\end{subfigure}
\begin{subfigure}{.13\textwidth}
  \centering
  \includegraphics[width=\textwidth]{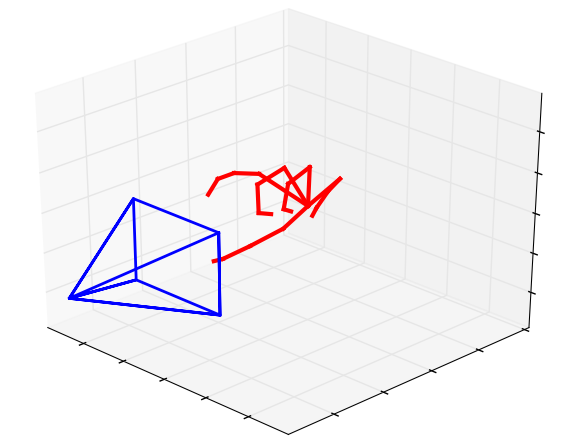}
\end{subfigure}
\begin{subfigure}{.13\textwidth}
  \centering
  \includegraphics[width=\textwidth]{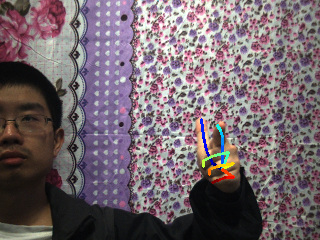}
\end{subfigure}
\begin{subfigure}{.13\textwidth}
  \centering
  \includegraphics[width=\textwidth]{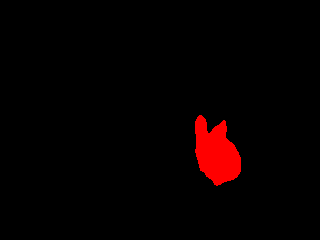}
\end{subfigure}
\begin{subfigure}{.13\textwidth}
  \centering
  \includegraphics[width=\textwidth]{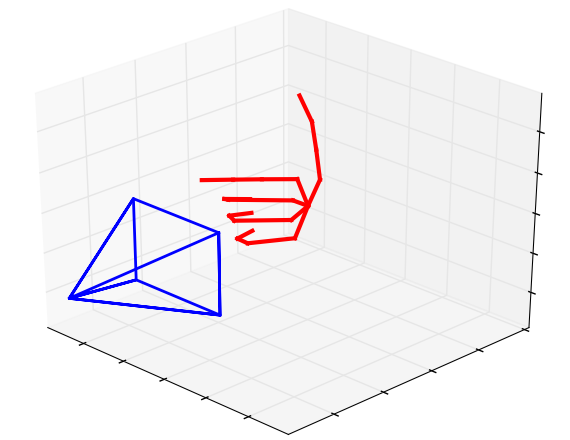}
\end{subfigure}

\begin{subfigure}{.13\textwidth}
  \centering
  \includegraphics[width=\textwidth]{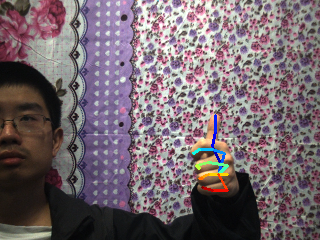}
\end{subfigure}
\begin{subfigure}{.13\textwidth}
  \centering
  \includegraphics[width=\textwidth]{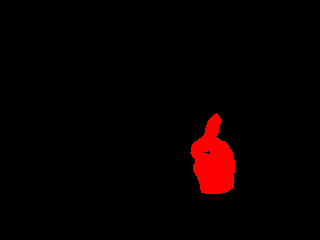}
\end{subfigure}
\begin{subfigure}{.13\textwidth}
  \centering
  \includegraphics[width=\textwidth]{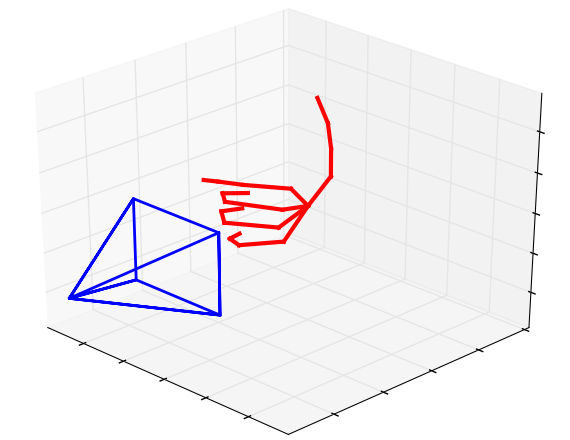}
\end{subfigure}
\begin{subfigure}{.13\textwidth}
  \centering
  \includegraphics[width=\textwidth]{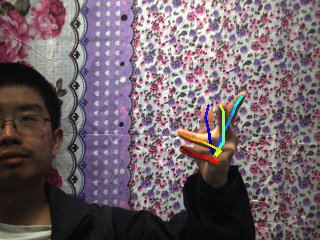}
\end{subfigure}
\begin{subfigure}{.13\textwidth}
  \centering
  \includegraphics[width=\textwidth]{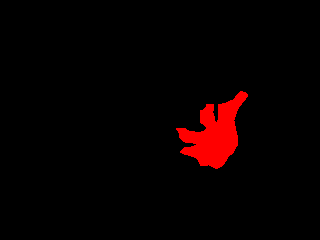}
\end{subfigure}
\begin{subfigure}{.13\textwidth}
  \centering
  \includegraphics[width=\textwidth]{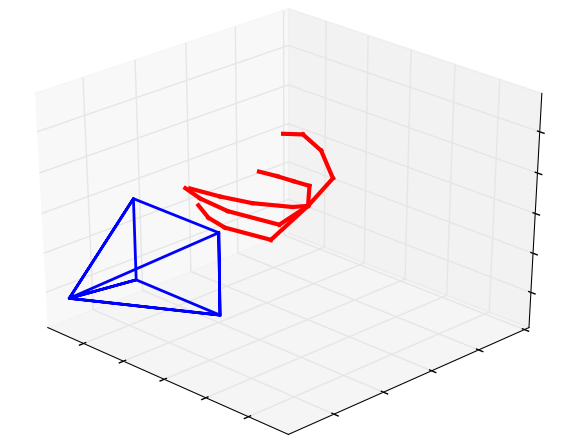}
\end{subfigure}
\caption{Qualitative examples of our complete system. Input to the network are color image and information if its a left or right hand.
The network estimates the hand segmentation mask, localizes keypoints in 2D (shown overlayed with the input image) and outputs the most likely 3D pose.
The top row shows samples from a dataset we recorded for qualitative evaluation, the following three rows are from \textit{R-val} and last three rows are from \textit{S-val}.}
\label{fig:results3dsecond}
\end{figure*}

\section{Additional results}
\label{sup:additional_results}
Figure \ref{fig:results3dsecond} shows results of the proposed approach.

\section{NYU Hand Pose Dataset}

\begin{figure}
\centering
\begin{subfigure}{.42\columnwidth}
  \centering
  \includegraphics[width=\textwidth]{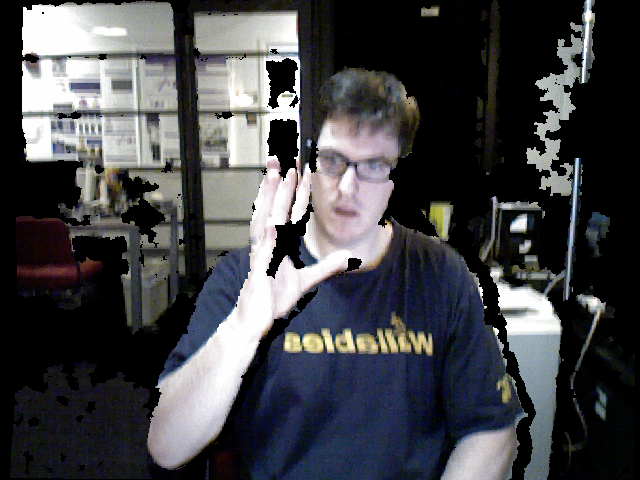}
\end{subfigure}
\begin{subfigure}{.42\columnwidth}
  \centering
  \includegraphics[width=\textwidth]{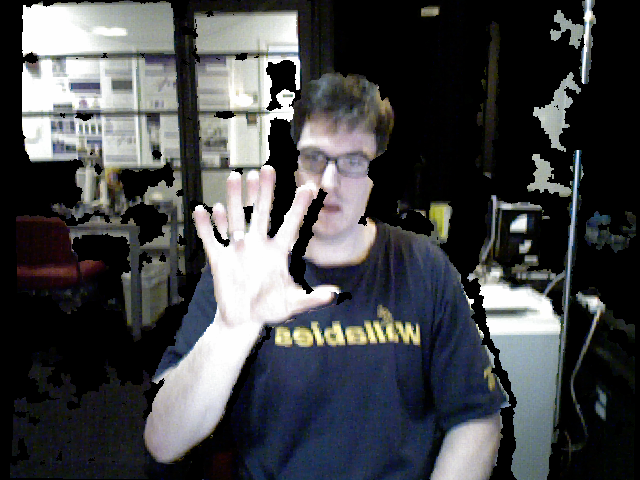}
\end{subfigure}
\caption{Two samples from the NYU Hand Pose Dataset by Tompson \etal~\cite{tompson_real-time_2014}. 
Due to artefacts in the color images this dataset is not suited to evaluate color based approaches.}
\label{fig:nyu_unsuited}
\end{figure}
 
A commonly used benchmark for 3D hand pose estimation is the NYU Hand Pose Dataset by Tompson \etal~\cite{tompson_real-time_2014}.
We can't use it for our work, because it only provides registered color images, which exclusively provide color information for pixels with valid depth data.
This results into corrupted images as shown in Figure~\ref{fig:nyu_unsuited}. This makes it infeasible to use for an approach that only utilizes color.

\end{document}